# Nested Junction Trees

Uffe Kjærulff
Department of Computer Science
Aalborg University
Fredrik Bajers Vej 7E, DK-9220 Aalborg Ø, Denmark
uk@cs.auc.dk

## Abstract

The efficiency of inference in both the Hugin and, most notably, the Shafer-Shenoy architectures can be improved by exploiting the independence relations induced by the incoming messages of a clique. That is, the message to be sent from a clique can be computed via a factorization of the clique potential in the form of a junction tree. In this paper we show that by exploiting such nested junction trees in the computation of messages both space *and* time costs of the conventional propagation methods may be reduced. The paper presents a structured way of exploiting the nested junction trees technique to achieve such reductions. The usefulness of the method is emphasized through a thorough empirical evaluation involving ten large real-world Bayesian networks and the Hugin inference algorithm.

## 1 INTRODUCTION

Inference in Bayesian networks can be formulated as message passing in a junction tree corresponding to the network (Jensen, Lauritzen & Olesen 1990, Shafer & Shenoy 1990). More precisely, a posterior probability distribution for a particular variable can be computed by sending messages inward from the leaves of the tree toward the clique (root) containing the variable of interest. If a subsequent outward propagation of messages from the root toward the leaves is performed, all cliques will then contain the correct posterior distributions (at least up to a normalizing constant). In many situations, however, we are only interested in the posterior distribution(s) for one or a few variables, which makes the outward pass redundant.

The Hugin and the Shafer-Shenoy propagation methods will be reviewed briefly in the following; for more in-depth presentations, see above references. We shall assume that all variables of a Bayesian network are discrete.

A Bayesian network consists of an independence graph, $G = (V, E)$, (which is an acyclic, directed graph or, more generally, a chain graph) and a probability function, $p$, which factorizes according to $G$. That is,

$$p_V = \prod_{v \in V} p(v | \mathrm{pa}(v)),$$

where $\mathrm{pa}(v)$ denotes the parents of $v$ (i.e., the set of vertices of $G$ from which there are directed links to $v$). The junction tree corresponding to $G$ is constructed via the operations of moralization and triangulation such that the nodes of the junction tree correspond to the cliques (i.e., maximal complete subgraphs) of the triangulated graph. To each clique, $C$, and each separator, $S$, (i.e., link between a pair of neighbouring cliques of the junction tree) is associated potential tables $\phi_C$ and $\phi_S$, respectively, by which, at any time, we shall denote the current potentials associated with $C$ and $S$.

Now define

$$\psi_C = \prod_i \phi_{V_i},$$

where $\phi_{V_i} = p(v_i | V_i \setminus \{v_i\})$ and $\mathrm{pa}(v_i) = V_i \setminus \{v_i\}$. That is, for each clique, $C$, is associated a subset of the conditional probabilities specified for the Bayesian network, and the function $\psi_C$ represents the product over this subset. Initially the potentials of the junction tree is given as

$$\phi_C = \psi_C \qquad \text{and} \qquad \phi_S = 1$$

for each clique, $C$, and each separator, $S$.

Propagation is based on the operation of absorption. Assume that clique $C$ is absorbing from neighbouring cliques $C_1, \ldots, C_n$ via separators $S_1, \ldots, S_n$. In the two architectures, this is done as indicated in Table 1.

|   Hugin  |  Shafer-Shenoy  |
|---|---|

Hugin:

1. $\phi^*_{S_i} = \sum_{C_i \setminus S_i} \phi_{C_i}, \; i = 1, \ldots, n$

2. $\phi^*_C = \phi_C \prod_{i=1}^{n} \dfrac{\phi^*_{S_i}}{\phi_{S_i}}$

3. $\phi_{S_i} := \phi^*_{S_i}, \; i = 1, \ldots, n$

4. $\phi_C := \phi^*_C$

Shafer-Shenoy:

1. $\phi^*_{S_i} = \sum_{C_i \setminus S_i} \phi_{C_i}, \; i = 1, \ldots, n$

2. $\phi^*_C = \psi_C \prod_{i=1}^{n} \phi^*_{S_i}$

3. $\phi_{S_i} := \phi^*_{S_i}, \; i = 1, \ldots, n$

Table 1: Absorption in the Hugin and the Shafer-Shenoy architectures.

Propagation can be defined as a sequence of inward message absorptions followed by a sequence of outward message absorptions, where inward means from leaf cliques of the junction tree towards a root clique, and outward means from the root clique towards the leaves. Note that the $\phi^*_{S_i}$'s are called *messages*. In the inward pass, since then $\phi_S = 1$ for all separators, the only difference between the two architectures is step 4 of the Hugin procedure (see Table 1).

In the outward pass, on the other hand, the difference between the two architectures becomes more pronounced. Consider clique $C$ which, in the inward pass, has absorbed messages from $C_2, \ldots, C_n$ and sent a message to $C_1$. Now, having received a message from $C_1$ in the outward pass, it is going to send messages to $C_2, \ldots, C_n$. In the two architectures, this is done as indicated in Table 2.

Note that in the Hugin architecture, when a clique $C$ absorbs a message $\phi^*_{S_i}$, it is always true that

$$\phi_{S_i} = \sum_{C \setminus S_i} \phi_C.$$

This fact is exploited in the Hugin architecture to avoid performing repeated multiplications. Hence, in the outward pass of the Hugin algorithm a clique $C$ can compute the product of all messages from its neighbours simply by 'substituting' one term of $\phi_C$ using division. Thus, the main difference between the Hugin and the Shafer-Shenoy architectures is the use of division. As we shall see later, avoiding division is advantageous when we use nested junction trees for inference.

The computation of messages is carried out as indicated in Tables 1 and 2, namely by multiplying all $\phi_{V_i}$'s and $\phi^*_{S_j}$'s together and marginalizing from that product. However, often $\phi^*_S$ can be computed via a series of marginalizations over smaller tables, which can greatly reduce both space and time complexities.

As a small illustrative example, assume that clique $C$ contains variables $\{X_1, X_2, X_3, X_4\}$, that $C$ receives messages $\phi_{\{X_1, X_2\}}$ and $\phi_{\{X_2, X_3\}}$, and that the potential $\phi_{\{X_3, X_4\}}$ was initially associated with $C$. The message, $\phi_{\{X_1, X_4\}}$, to be sent to $D$ is thus

$$\begin{aligned}
\phi_{\{X_1, X_4\}} &= \sum_{\{X_2, X_3\}} \phi_{\{X_1, X_2, X_3, X_4\}} \\
&= \sum_{\{X_2, X_3\}} \phi_{\{X_1, X_2\}} \phi_{\{X_2, X_3\}} \phi_{\{X_3, X_4\}} \quad (1)
\end{aligned}$$

However, since $\phi_{\{X_1, X_2\}}$ does not depend on $X_3$, we can compute $\phi_{\{X_1, X_4\}}$ as

$$\phi_{\{X_1, X_4\}} = \sum_{X_2} \phi_{\{X_1, X_2\}} \sum_{X_3} \phi_{\{X_2, X_3\}} \phi_{\{X_3, X_4\}}, \quad (2)$$

which reduces both space and time complexities: assuming all binary variables, Eq. 1 implies a space cost of 16 and a time cost (i.e., number of arithmetic operations) of 64 (3 × 16 for the multiplications and 16 for the marginalization), whereas Eq. 2 implies a space cost of 8 and a time cost of 48.

Basically, the trick in Eq. 2 is all there is to inference in Bayesian networks. In fact, the first general inference methods for probabilistic networks developed by Cannings, Thompson & Skolnick (1976) used exactly that method. Their method is referred to as "peeling", since the variables are peeled off one by one until the desired marginal has been computed. In most inference methods for Bayesian networks, finding the peeling order (elimination order) is done as an off-line, one-off process. That is, the acyclic, directed graph of the Bayesian network is moralized and triangulated (Lauritzen & Spiegelhalter 1988), and a secondary structure referred to as a junction tree (Jensen 1988) is constructed once and for all. The junction tree is then used as an efficient and versatile computational device.

Now, since Eq. 2 expresses nothing but inference in a probabilistic network consisting of four variables and three probability potentials, the computation of the message, $\phi_{\{X_1,X_4\}}$, can be formulated as inference in a junction tree with cliques $\{X_1, X_2\}$, $\{X_2, X_3\}$, and $\{X_3, X_4\}$. Thus, we have a junction tree in the clique of another junction tree! For slightly more complicated examples the nesting level might even be larger than two as shall be exemplified in Section 2, where we describe the construction of nested junction trees.

|  Hugin | Shafer-Shenoy |
|---|---|
| 1. $\phi_C^* = \phi_C \cdot \dfrac{\phi_{S_1}^*}{\phi_{S_1}}$ | 1. For $j = 2$ to $n$ do $$\phi_C^{*j} = \psi_C \cdot \prod_{i=1,\ldots,j-1,j+1,\ldots,n} \phi_{S_i}^*$$ $$\phi_{S_j}^* = \sum_{C \setminus S_j} \phi_C^{*j}$$ |
| 2. For $j = 2$ to $n$ do $$\phi_{S_j}^* = \sum_{C \setminus S_j} \phi_C^*$$ | |
| 3. $\phi_{S_1} := \phi_{S_1}^*$ | 2. $\phi_{S_1} := \phi_{S_1}^*$ |
| 4. $\phi_C := \phi_C^*$ | |

Table 2: Clique $C$ absorbs a message from clique $C_1$ and sends messages to its remaining neighbours.

Section 3 describes the space and time costs associated with computation in nested junction trees, and Section 4 briefly explains how the space and time costs of an inward probability propagation can be computed through propagation of costs. Section 5 presents the results of an empirical study of the usefulness of nested junction trees. Finally, in Section 6, we conclude the work by discussing the benefits and as well as the limitations of nested junction trees.

## 2 CONSTRUCTING NESTED JUNCTION TREES

To illustrate the process of constructing nested junction trees, we shall consider the situation where clique $C_{16}$ is going to send a message to clique $C_{13}$ in the junction tree of a subnet, here called Munin1, of the Munin network (Andreassen, Jensen, Andersen, Falck, Kjærulff, Woldbye, Sørensen, Rosenfalck & Jensen 1989). Clique $C_{16}$ and its neighbours are shown in Figure 1. For simplicity, the variables of $C_{16}$, $\{22, 26, 83, 84, 94, 95, 97, 164, 168\}$, are named corresponding to their node identifiers in the network, and they have 4, 5, 5, 5, 5, 5, 5, 7, and 6 states, respectively.

The set of probability potentials for a Bayesian network defines the cliques of the moral graph derived from the acyclic, directed graph associated with the network (notice that the directed graph is also defined by the potentials). That is, each potential $\phi_V$ (e.g., given by a conditional probability table) induces

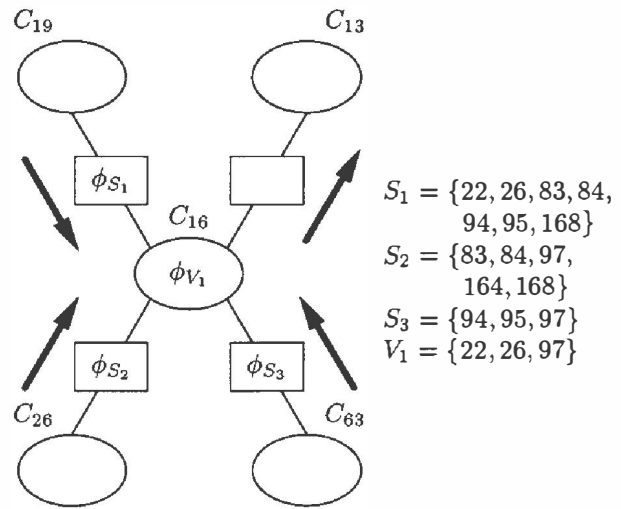

Figure 1: Clique $C_{16}$ receives messages $\phi_{S_1}$, $\phi_{S_2}$, and $\phi_{S_3}$ from cliques $C_{19}$, $C_{26}$, and $C_{63}$, respectively. Based on these messages and the probability potential, $\phi_{V_1} = P(97 | 22, 26)$, a message must be generated and sent to clique $C_{13}$.

a complete subgraph. A junction tree is then constructed through triangulation of the moral graph.

Thus, in our example, the undirected graph induced by the potentials $\phi_{S_1}$, $\phi_{S_2}$, $\phi_{S_3}$, and $\phi_{V_1}$ may be depicted as in Figure 2. At first sight this graph looks quite messy, and it might be hard to believe that its triangulated graph will be anything but a complete graph. However, a closer examination reveals that the graph is already triangulated and that its cliques are $\{83, 84, 97, 164, 168\}$ and $\{22, 26, 83, 84, 94, 95, 97, 168\}$.

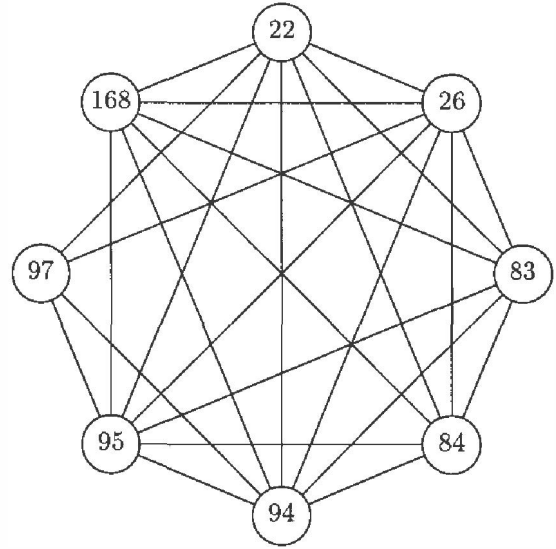

Figure 3: The undirected graph induced by potentials $\phi_{S_1}$, $\phi_{S_3}$, and $\phi_{V_1}$.

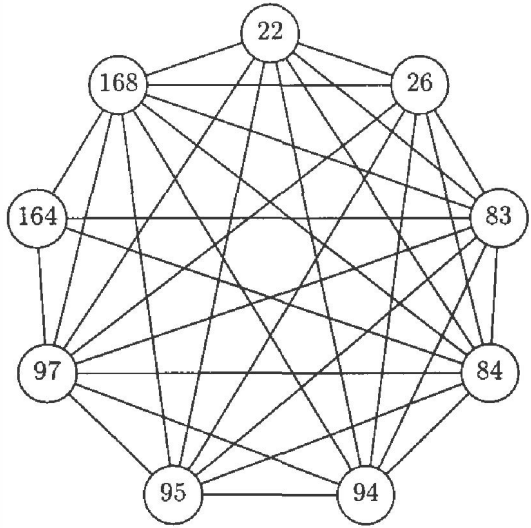

Figure 2: The undirected graph induced by potentials $\phi_{S_1}$, $\phi_{S_2}$, $\phi_{S_3}$, and $\phi_{V_1}$.

So, the original 9-clique (i.e., clique containing nine variables) with a table of size $2,625,000$ has been reduced to a junction tree with a 5-clique and an 8-clique with tables of total size $381,000$ (including a separator table of size $750$).

Thus encouraged we shall try to continue our clique break-down! In the two-clique junction tree, the 5-clique has associated with it only potential $\phi_{S_2}$, so it cannot be further broken down. The 8-clique, on the other hand, has got the remaining three potentials associated with it. These potentials (i.e., $\phi_{S_1}$, $\phi_{S_3}$, and $\phi_{V_1}$) induce the graph shown in Figure 3.

This graph also appears to be triangulated and contains the 5-clique $\{22, 26, 94, 95, 97\}$ and the 7-clique $\{22, 26, 83, 84, 94, 95, 168\}$ with tables of total size $78,000$ (including a separator table of size $500$). The reduced space cost is $375,000 - 78,000 = 297,000$.

In this junction tree, the 7-clique cannot be further broken down since it contains only one potential. The 5-clique, however, contains two potentials, $\phi_{S_3}$ and $\phi_{V_1}$, and can therefore possibly be further broken down. The two potentials induce the graph shown in Figure 4, hence a further break-down is possible as the graph is triangulated and contains two cliques.

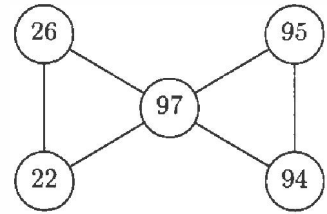

Figure 4: The undirected graph induced by potentials $\phi_{S_3}$ and $\phi_{V_1}$.

Now, no further break-down is possible. The resulting nested junction tree, shown in Figure 5, has a total space cost of $81,730$, which is significantly less than the original $2,625,000$. Carrying out the nesting to this depth, however, have a big time cost, since, for example, 500 message passings is needed through the separator $\{83, 84, 97, 168\}$ in order to generate the message from $C_{16}$ to $C_{13}$. A proper balance between space and time costs will most often be of interest. We shall address that issue in the next section.

## 3 SPACE AND TIME COSTS

As already discussed in Section 2, the smallest space cost of sending a message from a clique $C$ equals the accumulated size of the clique and separator tables of the nested junction tree(s) induced by the potentials of $C$ (i.e., messages sent to $C$ and potentials initially associated with it). For example, sending a message from clique $C_b$ (see Figure 5) has a smallest space cost of $75,000 + 500 + (100 + 5 + 125) = 75,730$. Note that this smallest space cost results from choosing clique $C_c$ as root. Choosing clique $C_d$ as root instead would make the inner-most junction tree (cliques $C_e$ and $C_f$) collapse to a single clique with a table of size 2,500, resulting in an overall space cost of 77,500. A similar analysis shows that, depending on which cliques are selected as roots, the space cost of generating the message for clique $C_{13}$ varies from 81,730 to 381,000.

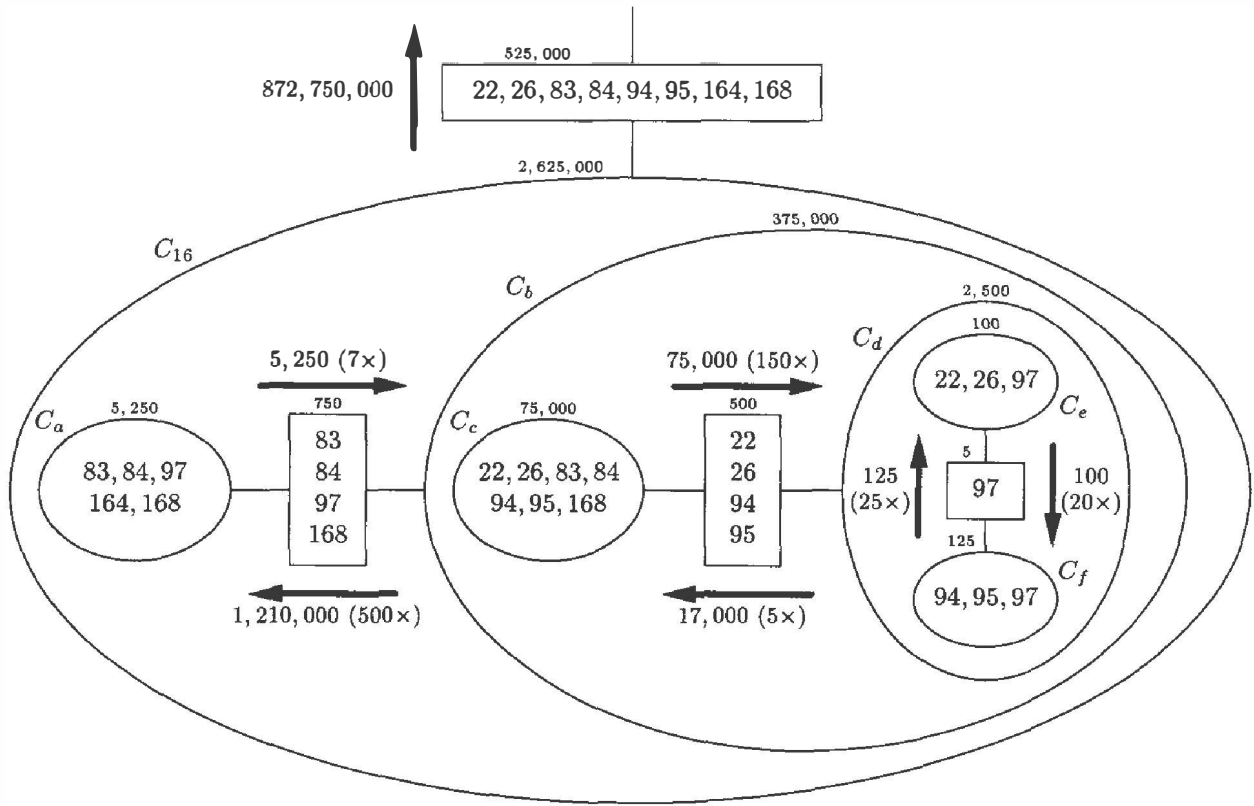

Figure 5: The nested junction tree for clique $C_{16}$ in Munin1. Only the connection to neighbour $C_{13}$ is shown. The small figures on top of the cliques and separators indicate table sizes, assuming no nesting. The labels attached to the arrows indicate (1) the time cost of sending a single message, and (2) the number of messages required to compute the separator marginal one nesting level up.

Let us consider the time cost of sending a single message from clique $C_b$ to clique $C_a$, and assume that clique $C_c$ is selected as root. To generate a message from $C_b$, $C_c$ must receive five messages from $C_d$, corresponding to the number of states of variable 97 which is a member of the $(C_a, C_b)$-separator but not a member of $C_c$. The time cost of each message from $C_d$ is 17,000 as shall be explained shortly.

Given this information we can now do our calculations. Generating a message involves the operations of multiplication (of the relevant potentials) and marginalization. The cost of multiplying the two potentials onto the clique table is 2 times the table size. The cost of marginalization equals the table size. Therefore, the time cost of sending a message from clique $C_b$ is

$$5 \times (17,000 + 3 \times 75,000) = 1,210,000.$$

Notice that the cost of marginalization equals the size of the larger of the clique table and the separator table. For example, whenever $C_a$ has received a message from $C_b$, it must basically run through the $(C_{16}, C_{13})$-separator table. Smart indexing procedures might, however, reduce that cost dramatically. So, our cost considerations given here are worst-case.

The time cost of 17,000 for sending a single message from $C_d$ to $C_c$ is found when selecting $C_f$ as root; selecting $C_e$ as root instead would have a cost of 20,625. Clique $C_e$ must send 20 messages for $C_f$ to be able to generate a single message to $C_c$. Notice, again, that this could be done much more efficiently, since for each message variable 97 is fixed, effectively splitting the inner-most junction tree into two. However, to keep the exposition as clear and general as possible, we shall refrain from introducing smart special-case procedures. Now using the same line of reasoning as above, we get the time cost of

$$20 \times (100 + 2 \times 125 + 500) = 17,000,$$

where each marginalization has a cost of 500 since the table of $C_f$ is smaller than the $(C_c, C_d)$-separator table.

The space and time costs of conventional (i.e., non-nested) message generation would in the $C_{16}$-to-$C_{13}$ case be $2,625,000$ and $5 \times 2,625,000 = 13,125,000$, respectively. The similar costs in the four-level nesting case are 81,730 and 872,750,000, which are 32 times smaller and 67 times larger, respectively, than the conventional costs. A more satisfying result is obtained by avoiding the two inner-most nestings (i.e., collapsing cliques $C_c$, $C_d$, $C_d$, and $C_f$, which happens if instead of $C_a$ we let $C_b$ be root), in which case we get costs 381,000 and 14,211,750 with the time cost being only slightly larger than in the conventional case, but with a 7 times reduction in the space cost.

## 4  PROPAGATION OF COSTS

The calculation of the costs of performing inward probability propagation toward a root clique $C$ can be formulated elegantly through propagation of costs in the junction tree. Let, namely, each clique send a cost message (consisting of a space cost and a time cost) being the sum of the costs of sending a message (i.e, a probability potential) and the sum of the cost messages from its remaining neighbours. Then a cost message states the cost of letting the sender be root in the subtree containing the sender and the subtrees from which it received its messages. Thus, when $C$ has received cost messages from all of its neighbours, the overall cost of an inward probability propagation is given by the sum of its cost messages plus the cost of computing the $C$-marginal potential.

Now, if we perform an outward propagation of costs from $C$, we will subsequently be able to compute the cost of an inward probability propagation to any other clique, just as we did for clique $C$!

## 5  EXPERIMENTS

To investigate the practical relevance of nested junction trees, the cost propagation scheme described above has been implemented as an extension to the Hugin algorithm. In order to find a proper balance between space and time costs, the algorithm makes a junction tree representation of a clique only if

$$\text{space\_cost} + \gamma \cdot \text{time\_cost},$$

is smaller than it is using conventional representation. The time factor, $\gamma$, is chosen by the user.

Cost measurements have been made on the following ten large real-world networks. The KK network is an early prototype model for growing barley. The Link network is a version of the LQT pedigree by Professor Brian Suarez extended for linkage analysis (Jensen & Kong 1996). The Pathfinder network is a tool for diagnosing lymph node diseases (Heckerman, Horvitz &

Nathwani 1992). The Pignet network is a small subnet of a pedigree of breeding pigs. The Diabetes network is a time-sliced network for determining optimal insulin dose adjustments (Andreassen, Hovorka, Benn, Olesen & Carson 1991). The Munin1-4 networks are different subnets of the Munin system (Andreassen et al. 1989). The Water network is a time-sliced model of the biological processes of a water treatment plant (Jensen, Kjærulff, Olesen & Pedersen 1989).

The average space and time costs of performing an inward probability propagation is measured for each of these ten networks. Table 5 summarizes the results obtained. All space/time figures should be read as millions. The first pair of space/time columns lists the costs associated with conventional junction tree propagation. The remaining three pairs of space/time columns show, respectively, the least possible space cost with its associated time cost, the costs corresponding to the highest average relative saving, and the least possible time cost with its associated space cost. The percentages in parentheses indicate the relative savings calculated from the exact costs. The highest average relative savings were found by running the algorithm with various $\gamma$-values for each network. The optimal value, $\gamma^*$, varied from 0.25 to 0.45.

Table 3 shows that the time costs associated with minimum space costs are much larger than the time costs of conventional (inward) propagation. Thus, although maximum nesting yields minimum space cost, it is not recommended in general, since the associated time cost may be unacceptably large.

However, as the $\gamma = \gamma^*$ columns show, a moderate increase in the space costs tremendously reduces the time costs. (The example in Figure 5 demonstrates the dramatic effect on the time cost as the degree of nesting is varied.) In fact, the time costs of conventional and nested computation are roughly identical for $\gamma = \gamma^*$, while space costs are still significantly reduced for most of the networks.

Interestingly, even though the time measures were absolutely worst-case, for all networks but Pathfinder the minimum time costs ($\gamma = 100$) are less than the time costs of conventional propagation, and, of course, the associated space costs are also less than in the conventional case, since the saving on the time side is due to nesting which inevitably reduces the space cost.

## 6  CONCLUDING REMARKS

The peeling inference method might exploit some of the extra independence relations available during inward probability propagation, and hence have space and time costs less than the conventional junction tree method for inward probability propagation, as indicated in Section 1. However, in the example shown in Figure 5 the peeling method is not able to exploit e.g. the conditional independence of variable 164 of variables 22, 26, 94, and 95 given variables 83, 84, 97, and 168. So, the technique presented in this paper is much more general than peeling.

Note that if the triangulated version of the graph induced by the separators of a clique is not complete (i.e., contains more than one clique), then one or more of the fill-in links of that clique are redundant; that is, the clique can be split into two or more cliques. Therefore, assuming triangulations without redundant fill-ins, the nested junction trees technique cannot be exploited in the outward pass of the Hugin algorithm, since messages have been received from all neighbours (including the recipient of the message). In the Shafer-Shenoy algorithm, on the hand, there is no difference between the inward and the outward passes, which makes the nested junction trees technique well-suited for that algorithm. A detailed comparison study should be conducted to establish the relative efficiency of the nested junction trees technique in the two architectures.

### Acknowledgements

I wish to thank Steffen L. Lauritzen for suggesting the cost propagation scheme, Claus S. Jensen for providing the Link and Pignet networks, David Heckerman for providing the Pathfinder network, Kristian G. Olesen for providing the Munin networks, and Steen Andreassen for providing the Diabetes network.

| | Conventional | | Nested | | | | | | |
| --- | --- | --- | --- | --- | --- | --- | --- | --- | --- |
| | | | $\gamma = 0$ | | $\gamma = \gamma^*$ | | | $\gamma = 100$ | |
| Network | Space | Time | Space | Time | Space | Time | | Space | Time |
| KK | 14.0 | 50.6 | 1.7 (88%) | 421.0 (−733%) | 3.5 (75%) | 33.9 | (33%) | 8.3 (41%) | 35.4 (30%) |
| Link | 25.7 | 83.3 | 2.4 (91%) | 39346.5 (−47135%) | 9.1 (65%) | 74.4 | (11%) | 17.9 (30%) | 72.8 (13%) |
| Pathfinder | 0.2 | 0.6 | 0.1 (31%) | 1.3 (−103%) | 0.2 (12%) | 0.7 | (−6%) | 0.2 (0%) | 0.6 (0%) |
| Pignet | 0.7 | 2.2 | 0.2 (75%) | 52.4 (−2295%) | 0.3 (54%) | 2.5 (−12%) | | 0.6 (12%) | 2.1 (3%) |
| Diabetes | 10.4 | 33.1 | 1.1 (90%) | 87.4 (−164%) | 1.1 (89%) | 42.3 (−28%) | | 9.8 (6%) | 31.5 (5%) |
| Munin1 | 188.4 | 729.9 | 29.2 (84%) | 384497.7 (−52577%) | 69.2 (63%) | 631.8 | (13%) | 122.8 (35%) | 595.2 (18%) |
| Munin2 | 2.8 | 9.7 | 0.7 (76%) | 220.9 (−2172%) | 1.4 (49%) | 11.0 (−13%) | | 2.6 (7%) | 9.3 (4%) |
| Munin3 | 3.2 | 12.1 | 0.6 (83%) | 225.8 (−1765%) | 1.4 (58%) | 13.3 (−10%) | | 2.4 (26%) | 12.0 (1%) |
| Munin4 | 16.4 | 64.3 | 5.4 (67%) | 470.0 (−631%) | 6.6 (60%) | 72.7 (−13%) | | 13.5 (18%) | 57.1 (11%) |
| Water | 8.0 | 28.7 | 1.0 (88%) | 2764.5 (−9528%) | 2.1 (74%) | 25.7 | (11%) | 2.7 (66%) | 25.5 (11%) |

Table 3: Space and time costs for inward propagation of probability potentials using (i) the conventional method, and the nested junction trees method with (ii) maximum nesting (minimum space cost), (iii) maximum average relative saving of space and time costs, and (iv) minimum time cost. The percentages in parentheses are the relative savings compared to conventional propagation.